\documentclass[runningheads,a4paper]{llncs}
\usepackage[dvips]{epsfig,color}
\usepackage{paralist} 

\long\def\comment#1{}
\newcommand{\bi}{\begin{itemize}}
\newcommand{\ei}{\end{itemize}}
\newcommand{\bc}{\begin{center}}
\newcommand{\ec}{\end{center}}
\newcommand{\letterspace}{\hspace*{1mm}}
\newcommand{\mybox}[1]{\mbox{\letterspace #1 \letterspace }}
\newcommand{\btbl}{\begin{tabular}}
\newcommand{\etbl}{\end{tabular}}

\newtheorem{mydefinition}{{\bf Def.}}

\newtheorem{myproblem}{{\bf Problem.}}

\newcommand{\De}{\Delta_e}

\newcommand{\T}{\mbox{${\cal T}$}}
\newcommand{\OU}{O}

\newcommand{\Eisodos} {{\bf\small Input}}
\newcommand{\Exodos}{ {\bf\small Output}}
\newcommand{\Bproblem}{{\bf\small Required Bnode Matching Problem}}

\usepackage{url}
\urldef{\mails}\path|{kristi, tzitzik}@ics.forth.gr|

\begin{document}


\title{Tasks that Require, or can Benefit
    from,  Matching Blank Nodes\thanks{
        Alternative Title:
        We will Never Escape from Anonymity:
        Tasks that Require (or can Benefit from)   Matching Blank Nodes     }
        }

\author{Christina Lantzaki \and Yannis Tzitzikas}
\institute{Computer Science Department, University of Crete,\\
Institute of Computer Science, FORTH-ICS, GREECE\\
\mails} \maketitle

\begin{abstract}
In various domains and cases,
we observe the creation and usage of information elements
which are {\em unnamed}.
Such elements do not have a name, or may have a name
that is not externally referable
(usually meaningless and not persistent over time).
This paper discusses why we will never `escape'
from the problem of having to construct
mappings between such {\em unnamed elements}
in information systems.
Since unnamed elements nowadays  occur very often
in the framework of the Semantic Web
and Linked Data as blank nodes,
the paper describes
scenarios
that can benefit from  methods
that compute {\em mappings} between the unnamed elements.
For each scenario, the corresponding bnode matching
problem is formally defined.
Based on this analysis,
we try to reach to more a general formulation
of the problem,
which can be useful for guiding
the required technological advances.
To this end,
the paper finally discusses methods to  realize blank node matching,
the implementations that exist,
and identifies open issues and challenges.
\end{abstract}

\section{Introduction}

In various domains,
from programming languages, databases to knowledge representation,
we observe the creation and usage of information elements
which are {\em unnamed}.
Such elements do not have a name,
or have a local name
that
is
not
externally
referable.
Such local names are usually meaningless and not persistent over time.
In the context of RDF these {\em unnamed} elements are called {\em blank nodes}, or for short {\em bnodes}. In this paper we discuss why we will never ``escape"
from the problem of having to construct {\em mappings} between unnamed elements
in information systems.
We justify this thesis
by describing some  basic tasks or scenarios
(i.e. equivalence, entailment, differential storage, synchronization, integration)
each having a step that requires solving a bnode matching problem.
This analysis makes also evident why we cannot bypass the problem of mapping bnodes,
by
just assigning to them names (local or global).
Instead we
{\em have to} treat them as unnamed elements for carrying out correctly these tasks.

\begin{figure}[h!] 	
    \vspace*{-3mm}
	\centering{
		\fbox{
        \includegraphics[scale=0.35]{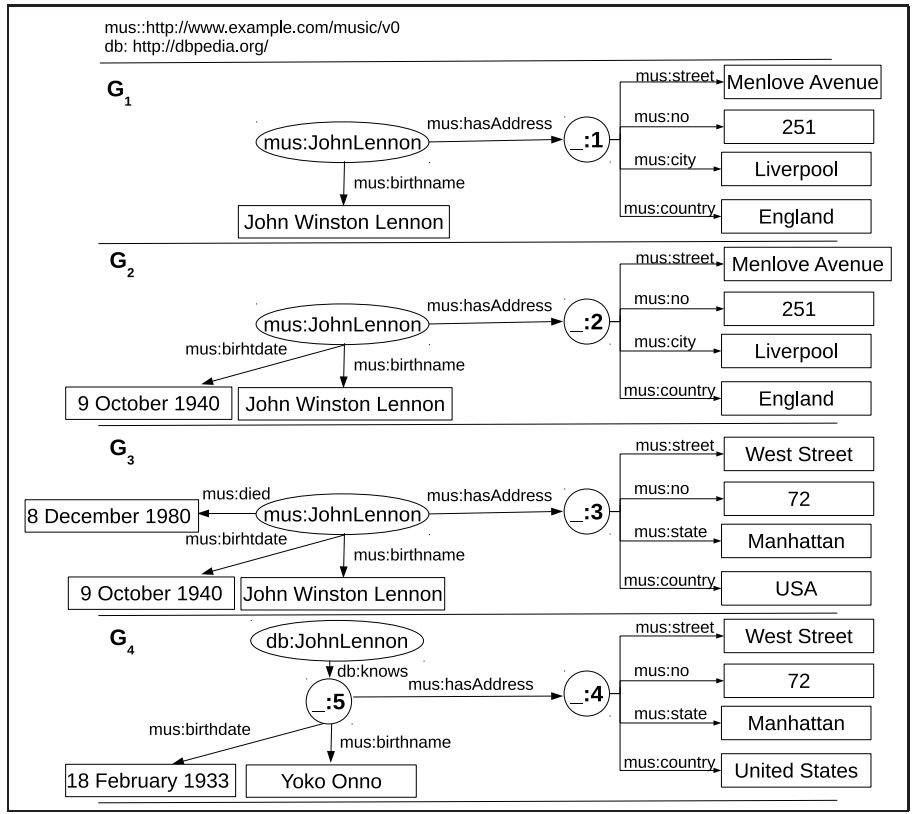}
        }
    	\caption{Running example}
		\label{fig:runExample}
    \vspace*{-3mm}
	}
\end{figure}

Since RDF is currently the ``lingua franca" for metadata,
and there is an increasing trend for publishing data
according to the principles of Linked Open Data (LOD), in the following scenarios
we consider RDF as the representation framework.
We should also note that a significant percentage of the resources
which are structured and/or published in the Semantic Web,
are unnamed \cite{mall-etal-2011ISWC}.
Figure \ref{fig:runExample} shows
a very simple running example comprising  four RDF graphs, $G_1$ to $G_4$.
All of them use classes and properties  defined in  a music ontology.
The first three graphs, $G_1, G_2$ and $G_3$, represent information about the musician
{\tt John Lennon}, while the fourth one, $G_4$, contains also information about {\tt Yoko Ono}.
We can see that {\tt John Lennon} is represented with two different URIs (i.e. {\tt mus:John$\_$Lennon}, {\tt db:John$\_$Lennon}) coming from different vocabularies.
We also observe five bnodes, where bnodes $\_\colon1$ to  $\_\colon4$ represent addresses (i.e. multi-component structures), while bnode $\_\colon5$ represents a person.
Notice also that $\_\colon4$ and $\_\colon5$ are directly connected. This running example will be used to describe the scenarios of Section \ref{sec:BScenarios}.

The key contributions of this paper are:
(a) it explains why we need to manage unnamed elements,
     i.e. why we will never escape from anonymity,
     by providing
      examples from various areas and technologies,
(b) it  provides concrete scenarios that require bnode matching
      and formulates  the required  matching problem,
(c) it consolidates all these problem formulations to a more general one,
and
(d) it describes what the current systems currently do, and identifies open
        issues and challenges.
The rest of this paper is organized as follows.
Section \ref{sec:Anonymous}
describes examples of unnamed elements in the broad sense
and their usual management.
Section \ref{sec:Background}
introduces the required background and notations.
Subsequently,
Section \ref{sec:BScenarios} describes
cases and scenarios
that can benefit from bnode matching.
Section \ref{sec:DC} discusses
methods that can be used
for realizing and offering bnode matching functionality,
and identifies
challenges and open problems.
Finally, Section \ref{sec:CR} concludes and identifies directions for further research.

\section{Anonymous Elements}
\label{sec:Anonymous}

\subsubsection{Examples of Anonymity.}
\label{sec:AnonymityGeneral}

Here we  provide examples of unnamed elements from  various domains.
Starting from our physical {\em universe} we can remark that it consists of unnamed elements:
atoms do not have any kind of external identity;
the identification of an atom  is  always done through its context.
The same is true for the molecules of living organisms: there is no external identity for these molecules; their identification is done through their context (i.e. whose body is this, in what part of the body they are located).
Also note that the human body itself is shedding old cells and is generating new ones being in a constant procedure of reconstruction.

Our {\em social life} also consists of unnamed elements.
It may be enough to identify a person through his/her first name within a group of friends, but this is not enough within a university. Although the registry number of a student is enough for identifying him/her within a university, that registry number is not enough within a country and so on.
Even when an entity is identified through a ``global" identifier
(e.g. a person through its passport number),
that identifier does not necessarily correspond to an intrinsic property of the entity,
and there is no guarantee that this identifier will always play this role
     (e.g.  passports can get lost, new passports are issued).

In a {\em Database Modeling} context,
specifically in ER (Entity-Relationship) Diagrams,
we have {\em weak entities},
i.e. entities which cannot be uniquely identified by their attributes alone.
To create a primary key for a weak  entity (and thus achieve unique identification),
we have to use a {\em foreign key} in conjunction with the entity's attributes.
The foreign key is typically a primary key of an (strong) entity which is related
to the (weak) entity that we want to identify.
Apart from ER Diagrams,
a relational table
quite commonly does not have any attribute
     that can serve as  global identity,
and in such cases keys are produced automatically by the DBMS.

Unnamed constructs also exist in {\em Programming Languages}.
For instance, in Java we have {\em unnamed classes},
    which are mainly used when building  graphical user interfaces.
    The compiler assigns to them a local name in the bytecode
    (the local name is derived by concatenating the name
        of the named class/interface that the anonymous class
        specializes/implements  and a counter).
In addition,
many times, {\em unnamed objects} are used  in parameter passing
(e.g. in a call like {\tt method(new Person())}).
Unnamed elements exist also in {\em collections},
e.g. in a list of lists, the list that is hosted as $x$-th element of the primary one,
does not have any identifier;
    it is identified through its position in the primary one.

RDF  supports unnamed elements, aka {\em blank nodes},
and several works (e.g. \cite{mall-etal-2011ISWC}) have demonstrated the usefulness of blank nodes for the representation of the Semantic Web data.
In a nutshell, from a theoretical perspective blank nodes play the role of the existential variables and from a technical perspective, as gathered in \cite{journals/jsw/ChenZCG12}, they give the capability to (a) describe multi-component structures, like the RDF {\em containers}, (b) describe {\em reification} (e.g. provenance information) and (c) represent {\em complex attributes} without having to name explicitly the auxiliary node (e.g. the address of a person consisting of the street, the number, the postal code and the city).

\cite{mall-etal-2011ISWC}  surveys the treatment of blank nodes in RDF data and proves the relatively high percentages of their usage. Indicatively, and according to the reported results,
the data fetched from the `rdfabout.com' and the `opencalais.com' domain, both of them parts of the LOD cloud, consist of 41.7$\%$  and 44.9$\%$ of blank nodes, respectively.
Finally, and in the general context of {\em Semantic Networks},
various methods have been proposed in the past
for assigning context-dependent names
(e.g. \cite{theodorakis1997context,theodorakis2002theory}).

\vspace*{-3mm}

\subsubsection{Naming/Identifying Unnamed Elements.}
\label{sec:UsualMgmt}

In many cases, local identifiers (i.e. not externally referable, trusted or persistent)
are assigned to the unnamed elements.
Cases that fit to this category are:
the autonumber keys in relational tables,
the automatic numbers assigned to unnamed classes in Java,
the blank node identifiers in an RDF triple of a file or a triplestore.
These identifiers are usually assigned to give the ability to refer more than once to the same element. However, the scope of the identifiers is strictly internal (e.g. the scope of an RDF file for the blank node identifiers).

In other cases, global identifiers are assigned to unnamed elements. In the context of RDF, URIs may be assigned to blank nodes through skolemization. However, this does not change the nature of the elements
in the sense  that even if we assign a global identifier to an element, permanency is not guaranteed.
Imagine a particular dataset (e.g. the {\tt DBpedia}  domain),
where the blank nodes are skolemized.
Nothing guarantees the reusability of the particular skolem for the description of the  same entity
by another dataset (e.g. the {\tt foaf} domain).

The key observation is that
{\em
\underline{we have to treat named elements as unnamed}
whenever we go out
of the intended  (systemic and/or time) scope
of the identification system
that assigned these names.
}
As consequence, even if we have named elements,
for certain tasks we have to treat them as unnamed!

\section{Background}
\label{sec:Background}

To describe precisely the scenarios that require (or can benefit from) bnode matching,
we need to introduce the necessary background and notations for RDF (in \S \ref{sec:RDFS}),
and
discuss the issue of URI matching
(in \S  \ref{sec:URIMatching})
and renaming (in \S \ref{sec:Renamings}).

\subsection{Preliminaries}
\label{sec:RDFS}

\subsubsection{RDF.}
Consider an infinite set $U$ (RDF URI references),
an  infinite set $B$ (blank nodes)
and an infinite set $L$ (literals).
A triple $(s, p, o) \in (U \cup B) \times U \times (U \cup B \cup L)$
is called an {\em RDF triple}
($s$ is called the {\em subject}, $p$ the {\em predicate}
and $o$ the {\em  object}).
Let $\T$ be the set of all possible triples,
i.e. $\T = (U \cup B) \times U \times (U \cup B \cup L)$.
An RDF Knowledge Base (KB) $K$,
or equivalently an {\em RDF graph} $G$,
is a finite set of RDF triples
(i.e. $K \subset \T$).
For an RDF graph $G_1$ we shall use $U_1, B_1, L_1$
to denote the URIs, bnodes and literals
that appear in the triples of $G_1$ respectively.
The {\em nodes} of $G_1$ are the values
that appear as subjects or objects
in the triples of $G_1$.
The equivalence of {\em RDF graphs}
is defined
in \cite{RDFSyntax} as:

\begin{mydefinition}[Equivalence of RDF Graphs that contain Bnodes] \label{def:standardEquiv}\  \\
\emph{Two RDF graphs $G_1$ and $G_2$ are {\em equivalent} if there is a
bijection\footnote{
    A function that is both one-to-one (injective) and onto (surjective).
} $M$ between the sets of nodes  of the two graphs ($N_1$ and $N_2$), such that:
\vspace*{-2mm}
\bi
\item $M(uri)=uri$ for each $uri$ $\in$ $U_1 \cap N_1$
\item $M(lit)=lit$ for each $lit$ $\in$ $L_1$
\item $M$ maps bnodes to bnodes (i.e. for each $b \in B_1$ it holds $M(b) \in B_2$)
\item A triple $( s, p, o )$ is in $G_1$ if and only if a triple
        $( M(s), p, M(o) )$ is in $G_2$. $\diamond$
\ei  }
\end{mydefinition}

\vspace*{-1mm}

It follows that if two graphs are equivalent then it certainly holds $U_1 = U_2$,
$L_1=L_2$ and  $|B_1|=|B_2|$.

\subsection{URI Matching}
\label{sec:URIMatching}

In general, in information comparison and/or integration
we need methods to compare and map URIs and literals
that come  from  different sources.
There are more than one methods, or policies, for doing so.
Below we distinguish three main policies for defining
a relation of {\em equivelance} over URIs
(based on  \cite{tzitzik2014connectivity}):

\noindent {\bf [i]}
 {\em Exact String Equality.}
    We treat two URIs $u_1$ and $u_2$ as equivalent,
    denoted by $u_1 \equiv_{[i]} u_2$,
    if $u_1 = u_2$  (i.e. strings equality).

\noindent {\bf [ii]}
{\em Suffix Canonicalization.}
    Here
    we treat two URIs $u_1$ and $u_2$ as equivalent,
     denoted by $u_1 \equiv_{[ii]} u_2$ if
    $last(u_1) = last(u_2)$
    where
    $last(u)$
    is the string obtained by
    (a) getting the substring after the last "/" or "\#", and
    (b) turning the letters of the picked substring  to lowercase
    and deleting the underscore letters that might exist.
    According to this policy
    \url{http://www.dbpedia.org/John_Lennon}
    $\equiv_{[ii]} $
    \url{http://www.example.com/music/v0/John_Lennon}
    since
    their  canonical suffix
    is the same, i.e. {\tt John$\_$Lennon}.

    This method in used in the warehouse described in \cite{tzitzik2013marineTLO}.
    However, this is just indicative in the sense that various other methods
    or distance functions (e.g. \cite{stoilos2005string}) could be used.
    With policy [ii] we actually want to refer to methods
    where   URI equivalence is based on a series or transformation and comparisons
    of the URI strings.

\noindent {\bf [iii]}
{\em Entity Matching.}
    Here consider
    $u_1 \equiv_{[iii]} u_2$ if
    $u_1$ {\tt sameAs} $u_2$
    according to an entity matching
    approach,
    i.e. the approach
    that could be eventually used
    in an information integration context.
    In general,
    an entity matching approach,
    independently of its internals
    (i.e. whether it is based on lexical and/or structural similarity
    and/or  other criteria or constraints),
    it
    results in the creation
    of {\tt sameAs} relationships between URIs.
    For instance,
    \cite{tzitzik2014connectivity}
    employs  SILK \cite{volz2009silk}\footnote{
        \small{\url{http://wifo5-03.informatik.uni-mannheim.de/bizer/silk/}}
    }
    for formulating and applying such rules
    over an operational semantic warehouse.

\ \\
Note that if two URIs are equivalent according to policy [i],
then they are equivalent according to [ii] too
(i.e. $\equiv_{[i]} \subseteq  \equiv_{[ii]}$).
Policy [i] is very strict (probably too strict for matching entities coming from different sources),
however it does not produce any false-positive.
Policy [ii] achieves treating as equal entities
across different namespaces,
however false-positives may occur.
For instance,
{\tt Argentina} is a {\em country}
(\url{http://www.fishbase.org/entity#Argentina})
but also a {\em fish genus}
(\url{http://www.marinespecies.org/entity#WoRMS:125885/Argentina}).
Policy [iii]  is fully aligned with the intended query behaviour
in an information integration context
but requires the application of an entity matching approach
(generic or domain specific).

\subsection{Renamings}
\label{sec:Renamings}

Let us  now introduce some additional notations for {\em replacement}
(or renamings).
{\em Resource} is any element of $U \cup B$ (for short $UB$).
Given a triple $t$ and
a pair of resources $(a,b)$
 we shall use the notation
 $t\#(a,b)$ to denote the triple
 obtained by {\em replacing} $a$ by $b$ if $a$ appears in the subject or/and object of $t$.
 If $a$ does not appear in $t$, then $t\#(a,b) = t $.
 For example,
 $(a,b,c)\#(c,d) = (a,b,d)$,
 while
 $(a,b,c)\#(e,f) = (a,b,c)$.
 Formally:
\bc
$t\#(a,b) =  \left\{ \begin{array}{ll}
                      (s,p,o)  ,  & \mybox{if}  t=(s,p,o), s\neq a, o\neq a \\
                      (b,p,o)  ,  & \mybox{if}  t=(a,p,o), o\neq a  \\
                      (s,p,b)  ,  & \mybox{if}  t=(s,p,a), s\neq a\\
                      (b,p,b)  ,  & \mybox{if}  t=(a,p,a)\\
                    \end{array}\right.
$
\ec

\noindent
We can generalize the notation and allow not only one pair of resources, like  $(a,b)$,
but  any {\em partial function} (hereafter just function) from $UB$ to $UB$.
In particular, if $F$ is a function ($F: UB \rightarrow UB$),
then $t\#F$
denotes the triple
 obtained by replacing every resource $a$ of $t$
 that belongs to the domain of $F$ by the image of $a$, i.e. by  $F(a)$.
This means that  if  $t=(a,p,o)$ and $a \in dom(F)$,
e.g. suppose that $F = \{(a,k),(w,z)\}$,
then $t\#F = t\#(a,f(a)) = (f(a),p,o)=(k,p,o)$.
 We can further generalize the notation
 for specifying replacements not only in one triple,
 but to a set of triples.
 Specifically, if  $T$ is a set of triples,
 then  $T\#R = \{~ t\#R ~|~ t\in T\}$.

\section{Tasks and Scenarios that Require, or can Benefit from  Bnode Matching}
\label{sec:BScenarios}

This section describes
a number of tasks and scenarios that require, or can benefit from,
bnode matching.
There are two different directions that are closely related with
bnode matching.
 The first direction is the field of {\em data management}. Equivalence checking, entailment checking, versioning and synchronization require a matching of the blank nodes in order to become more efficient. The second direction is the field of {\em data integration}. Instance matching (and by extension bnode matching) aims at matching
the same real world entities coming from different sources.

In brief,
the following tasks require blank node matching
either for solving  a decision (\S \ref{sec:Equivalence}, \S \ref{sec:Entailment})
or an optimization problem (\S \ref{sec:DeltaBasedStorage}, \S \ref{sec:synch}, \S \ref{sec:integration}).
We shall use
 $G_1$ and $G_2$ to denote  two RDF graphs. For simplicity reasons, and without loss of generality,
unless otherwise stated, below we will assume that $|B_1| \leq |B_2|$.

\subsection{Equivalence}
\label{sec:Equivalence}

The objective here it to decide whether two graphs are equivalent, i.e. whether $G_1 \equiv G_2$.
It is evident from the definition of equivalence (Def. \ref{def:standardEquiv}) that bnode matching is required.
Recall that the named resources (i.e. URIs) as well as the literals can be easily matched from the one graph to the other as identity functions. The complexity of the problem arises because of anonymous resources. Under the existence of blank nodes the problem is formulated as follows: Is there a bijection of the blank nodes in $G_1$ with the blank nodes in $G_2$ such that $G_1 \equiv G_2$ is satisfied?
\cite{Gutierrez:2004} proves that deciding equivalence of simple RDF graphs is Isomorphism-Complete. Note that the graph isomorphism problem belongs to the NP class (but not known to belong to NP-Complete).
The decision problem and the required bnode matching problem can be formulated as follows:

\begin{myproblem}[Testing Equivalence]\ \label{problem:Equivalence} \\
\Eisodos: $G_1, G_2$ \\
\Exodos:  $True$ if $G_1 \equiv G_2$, $False$ otherwise.\\
\Bproblem:\\
To check if $G_1 \equiv G_2$,
it requires checking if there exists
 a bijective function $M: B_1 \rightarrow B_2$
that satisfies the condition of the last bullet of Def. \ref{def:standardEquiv}.
\end{myproblem}

\noindent
As an example assume that we want to check equivalence between the graphs $G_1$ and $G_2$ of the running example. Matching bnodes $\_:1$ and $\_:2$ (i.e. $M$=$\{(\_:1,\_:2)\}$) gives that the triples $(\{(s,p,o) \in G_1 ~|~ s=\_:1 \} \cup \{(s,p,o) \in G_1 ~|~ o =\_:1\})$ $\#$ $M$ $\equiv$ $\{(s,p,o) \in G_2 ~|~s=$ $\_\colon2$ $\}$ $\cup \{(s,p,o) \in G_2 ~|~ o=\_:2\}$. However, the triple ($mus:JohnLennon$, $mus:birthdate$, $``9 October 1940"$) is in graph $G_2$, but not in $G_1$. Therefore, the answer to the question
$G_1 \equiv G_2$ is $False$.
If this triple was neither in $G_2$ (or was in both $G_1$ and $G_2$) the answer would be $True$. In both cases, bnode matching was necessary to decide equivalence.

\subsection{Entailment}
\label{sec:Entailment}

The objective of simple entailment is to find out if a graph $G_1$ entails a graph
$G_2$, i.e. whether $G_1 \models G_2$.
This problem returns a positive answer if and only if every interpretation that
satisfies $G_1$ also satisfies $G_2$.
In most scenarios simple entailment (i.e. RDF entailment) is restricted to vocabulary entailments (i.e. RDFS entailment) by restricting to a specific set of interpretations (i.e. rdfs-interpretations).
In \cite{Gutierrez:2004} it is proved that deciding simple or RDFS entailment of RDF graphs is NP-Complete.
The intractability of the problem depends only on blank nodes in the RDF graphs.
Let us focus on simple RDF entailment.
In case there are no blank nodes,
each triple is independent of the other and a simple existence test of the triples in $G_1$ is enough \cite{Gutierrez:2004} to give us if $G_2$ is a subgraph of $G_1$.
Under the existence of blank nodes in the graphs the problem is formulated as follows: Is there a function that maps the blank nodes of $G_2$ with the blank nodes or the URIs of $G_1$ such that $triples(G_2)\#m \subseteq	G_1$ is satisfied?
Note that a blank node of $G_2$ can be mapped to one or more blank nodes and Uris. It holds that this function is not necessarily injective or surjective.  Think of the case where
$G_1$ = $\{$(\url{http://Andre}, name, ``Andre"), (\url{http://Andre}, surname, ``Smith"), ($\_:1$, name, ``Natalie")$\}$
and $G_2$ = $\{$($\_:2$, name, ``Andre"), ($\_:3$, surname, ``Smith")$\}$.
For the map $m$ = $\{$($\_:2$, \url{http://Andre}), ($\_:3$, \url{http://Andre})$\}$, it holds that $triples(G_2)\#m \subseteq G_1$ and therefore $G_1 \ entails \ G_2$.
In case the two graphs do not contain directly connected blank nodes,
a polynomial solution is provided \cite{Gutierrez:2004}.
The method actually matches each blank node $b_2$  of $G_2$,
with all the blank nodes of $G_1$ and decides on the final injection.
If there is not such a function we get that $G_1$ does not entail $G_2$.
If blank nodes are directly connected, they cannot be tested independently and
the number of possible mappings becomes exponential in size.
However, authors in \cite{Gutierrez:2004} manage to prove that
the function between the blank nodes can be given in polynomial time
if $G_2$ contains blank node structures with bounded treewidth.

\begin{myproblem}[Testing Entailment]\
\label{problem:Entailment} \\
\Eisodos:  $G_1$, $G_2$  \\
\Exodos:  True if $G_1 \models G_2$. \\
\Bproblem: \\
To check if $G_1 \models G_2$,
it requires finding
a function $m: B_2 \rightarrow B_1 \cup U_1$ such that
    for each blank node $b_2$ of $G_2$,
    it holds:
    $triples(b_2)\#m \subseteq G_1$.
    We use $triples(b_2)$
    to denote the subset of triples in $G_2$ in which $b_2$ participates.
\end{myproblem}

\noindent
Making use of the running example we want to answer if $G_2$ entails $G_1$ (i.e. $G_2 \models G_1$). In order to test if $G_1$ is a sub-graph of $G_2$ bnode $\_\colon1$ is matched with bnode $\_\colon2$ (i.e. $m$=$\{(\_\colon1,\_\colon2)\}$). For this map 
we get that the triples $\{triples(\_\colon1 ) \in G_1\}\#$ $m$ $\subseteq$ $\{triples(\_\colon2) \in G_2\}$.
 So, with the help of bnode matching we got that the answer is $True$.

\subsection{$\Delta$-based Storage for Versioning}
\label{sec:DeltaBasedStorage}

Consider a versioning scenario
where we have a sequence of versions $G_1, G_2,\ldots, G_k$
(where $G_k$ is the latest version),
and we want to be able to fetch any past version, i.e. any $G_i$ for $1\leq i \leq k$.
Instead of having to store
the entire sequence
$\langle G_1,\ldots,G_k\rangle$,
a space economical way is to
adopt a {\em delta-based storage},
i.e. to  store  the initial (resp. last) version
and the {\em forward} (resp. {\em backwards}) deltas.
Let us focus on the forward deltas policy.
We store
$\langle G_1, \Delta(G_1 \rightarrow G_2),
\ldots,
\Delta(G_{k-1} \rightarrow G_k)\rangle $.
Although {\em delta-based storage} can reduce a lot the required space,
 the inability to match blank nodes
can keep unnecessarily high the delta size
 as recognized in \cite{zdiff07,mall-etal-2011ISWC}.
Bnode matching can reduce the delta size,
because after it is applied, triples that contain blank nodes may become the same and thus they are not reported in the delta.
Specifically, if we match a bnode $b_1 \in B_1$ to a bnode $b_2$ (of $B_2$),
    through an injection $M$,
    then these bnodes can be considered as equal at the computation of delta.
Formally, for the differential function $\De$, instead  of
computing $ \De(G_1 \rightarrow G_2) =  \{Add(t)~|~t\in G_2 \setminus G_1\} \cup \{Del(t)~|~t\in G_1 \setminus G_2\}
$,
the availability of an injection $M: B_1 \rightarrow B_2$,
allows
defining and computing:
\begin{equation} \label{eq:x}
\De(G_1 \rightarrow G_2, M) = \{Add(t)~|~t\in G_2 \setminus (G_1\#M) \} \cup \{Del(t)~|~t\in (G_1\#M)\setminus G_2\}
\end{equation}
which contains the same or less operations.
The operation $Add(t)$ denotes the addition of the triple $t$, while the operation $Del(t)$ denotes the deletion.
Notice that the resulting update operations
do not contain any bnode from $B_1$, since they have been replaced by bnodes in $B_2$
(due to $(G_1\#M)$ as defined in \S  \ref{sec:RDFS}).
If instead we want the resulting update operations
to contain only bnodes from $B_1$ (and none from $B_2$),
then we can define:
\begin{equation} \label{eq:forward}
\De(G_1 \rightarrow G_2, M) = \{Add(t)~|~t\in (G_2\#M) \setminus G_1 \} \cup \{Del(t)~|~t\in G_1 \setminus (G_2\#M)\}
\end{equation}
Having $G_1$ and $\De(G_1 \rightarrow G_2, M)$
we (the versioning system) can compute a graph that is equivalent to $G_2$.
The {\em backwards} deltas policy is also possible by using the mappings reversely.

\begin{myproblem}[Delta-based Storage]\ \label{problem:Delta} \\
\Eisodos:  $G_1, G_2$\\
\Exodos: A $\Delta(G_1 \rightarrow G_2)$ usable for delta-based storage in a versioning system\\
\Bproblem:
Find an injection $M: B_1 \rightarrow B_2$ that minimizes the delta of the unnamed parts of the
graphs. \\
\end{myproblem}

\noindent
As regards the problem of finding this injection,
\cite{bnodesDelta:2012}
proves that building a mapping between the blank nodes of two compared Knowledge Bases
that minimizes the delta size is NP-Hard in the general case,
and polynomial if they are not connected bnodes. That paper also
provides various approximation algorithms,
and various experimental results that quantify the
delta reduction gains.
\noindent
Returning to our running example, consider
that we have three different versions, $\langle$ $G_1$, $G_2$, $G_3$ $\rangle$,
 and we want to apply a {\em forward} {\em delta-based storage}.
 In such case we have to compute  $\De(G_1 \rightarrow G_2)$ and $\De(G_2 \rightarrow G_3)$. For the first comparison we have that the personal information of {\tt John Lennon} was enriched by his birthday. The mapping $M_1$ = $\{$($\_:1$, $\_:2$)$\}$ allows to match the address of $G_1$ to the address of $G_2$ and realize that the contained triples are the same. The output delta is $\De(G_1 \rightarrow G_2, M_1)$ =$\{Add(mus:John\_Lennon,mus:birthday, ``9 October 1940")\}$. Thanks to bnode matching the delta size is minimized from 11 operations to 1.
\noindent
For the second comparison we have that {\tt John Lennon} has changed his address. The bnode mapping is $M_2$ = $\{$($\_\colon2$, $\_:3$)$\}$ and matches the two addresses. However, none of the outgoing triples of $\_\colon$ are in $G_3$ and similarly none of the outgoing triples of $\_:3$ are in $G_2$. So, $\De(G_2 \rightarrow G_3, M_2)$ = $\{Del(\_:2,street, ``Menlove Avenue"), Del(\_:2,no,``251"), Del(\_:2,city,``Liverpool")\}$, $Del(\_:2,country,``England")$,
$Add(\_:2,street,``West$\\ $Street"), Add(\_:2,no,``72"),$ $ Add(\_:2,state,``Manhattan")\}$, $Add(\_:2,country,$ \\$``USA")$. Now, bnode matching minimized the delta size from 11 to 9 operations. The delta-based storage is achieved. If the versioning system is asked to fetch $G_2$ it can easily construct it by adding the triple $(mus:John\_Lennon,mus:birthday, ``9 October 1940")$ to $G_1$. Similarly, for the graph $G_3$.


\subsection{Synchronization}
\label{sec:synch}

Consider a server $S$ responsible for maintaining  a corpus of information
(e.g. information about public transport, tourism, events, etc)
that is represented in the form of an  RDF Graph $G_S$. Also consider several clients (e.g. persons having smart mobile devices) that  are interested in having this corpus of information
also stored at their side, for using it fast, even off line.
For this reason each client $C_i$
has subscribed to the server,
with the objective of  getting and having at its side
a graph $G_i$ such that $G_i \equiv G_S$.
Whenever the $G_S$ changes,
the client $C_i$ would like to get informed
and get the new version.

\noindent
To realize this scenario,
one straightforward policy is the following:
whenever $G_S$ changes, the server
sends the entire $G_S$ to the clients. A policy that requires sending less information can be based on deltas. Let $G_i$ denote the graph that is stored at the client after that step,
and clearly it holds $G_i \equiv G_S$.
Now suppose that $G_S$ changes and let $G_{S'}$ denote
the new graph.
Instead of sending the entire graph,
the server can compute and send to the client
the $\Delta(G_S \rightarrow G_{S'},M)$ as defined in \S \ref{sec:DeltaBasedStorage}.

The client receives this piece of information,
and then it has to apply the received operations on $G_i$
for getting a $G_{i'}$
such that $G_{i'} \equiv G_{S'}$.

\noindent
However, in order to ensure that the received operations from the server $S$ can be applied to the client $C_i$, the client should use a triple store system that enables persistence over the blank node identifiers. Applying persistence over both the received graph and the received deltas ensures that both parts share the same local names. Persistence in the server side is not necessary, as in case the blank node identifiers are changed, the server could compute the injection of the previous identifiers to the new ones and send this injection to the client. Applying this injection the client can keep up with the new identifiers.

\noindent
We should note that this task works with the exact same way as the $\Delta$-based Storage for Versioning. The only difference is the way the output is tackled (i.e. now the delta is sent to a client).

\subsection{Integration}
\label{sec:integration}

The objective here is to integrate data.
Consider a {\em materialized} (as opposed to the {\em mediator}) integration approach,
where data are fetched from several remote sources
and are placed
(in their original form
or after transformation)
in a single system, aka {\em warehouse},
which provides a unified access to this information.
For instance, suppose we want to aggregate the data
from the sources $G_1$ and $G_2$ of the running example (Figure \ref{fig:runExample}). Note that there is no schema discrepancy,
or assume that we have already tackled such discrepancies.
In other words, we are in one of the following cases:
(a)  both sources use the same schema (like in the running example),
(b) the sources use different schemas
but the fetched content is transformed  according to a common schema
when placed in the warehouse,
(c) the sources use different schema
but the warehouse contains schema mappings
(comprising  {\tt subClassOf, subPropertyOf,
owl:equivalentProperty,  owl:equivalentClass} properties, etc)
between their schemas
and the  warehouse schema
(e.g. as in the operational warehouse described in \cite{tzitzik2013marineTLO}).

Without bnode matching
we will consider
that {\tt John Lennon} has certainly two addresses.
Even if the property {\tt hasAddress} is not functional,
it is more reasonable to apply bnode matching
which in turn would return
the pair {\tt (\_:1, \_:2)}
as a matching pair.
Then the curator (or system policy)
can accept it,
or accept it after inspection.
Acceptance here
means
that either {\tt \_:1} and {\tt \_:2}
in the warehouse will be
assigned the same local name,  say {\tt \_:z},
or that a triple
{\tt (\_:1 owl:sameAs \_:2)}
will be added to the repository.

\begin{myproblem}[Integration]\ \label{prob:IntegrationFormulation} \\
We want to integrate information coming from two or more sources.\\
\Eisodos:  Two graphs $G_1$ and $G_2$ and
           an equivalence relation
           (extensionally or intentionally specified)
           between their
           resources, i.e. a relation
            $\equiv \ \subseteq
            (U_1 \cup B_1 \cup L_1) \times
            (U_2 \cup B_2 \cup L_2)$.
             The relation $\equiv$
    was  discussed
    in \S \ref{sec:URIMatching}.
            \\
\Exodos:  
    The input ($G_1, G_2, \equiv$)
    and a bnode mapping between $B_1$ and $B_2$.
 \\
\Bproblem:
    A relation between $B_1$ and $B_2$.
    In comparison to
    Problem \ref{problem:Delta}
    here  during the computation of the edit distances \cite{bnodesDelta:2012}
    (or signature equality \cite{bnodesDelta:2012}),
    we  should not consider string equality ($=$),
    but the equivalence ($\equiv$) that was provided as input.
\end{myproblem}

\noindent
Now consider the 4 graphs $G_1$ to $G_4$ of the running example.
By ignoring namespaces
(i.e. considering policy [ii] and policy [iii]
of \S \ref{sec:URIMatching}),
it follows that we have two persons:
{\tt John Lennon} ({\tt mus:John$\_$Lennon} $\equiv$ {\tt db:John$\_$Lennon}) and {\tt Yoko Ono}.
We have four bnodes representing addresses.
By comparing all sets of bnodes in pairs,
i.e.
$B_1$ with $B_2$,
$B_1$ with $B_3$, and so on,
we will come up
with pairs
of the form
{\tt  (\_:1, \_:2)},
{\tt  (\_:3, \_:4)}.
If we accept these then
the integrated information
will contain two persons, {\tt John Lennon} with two addresses one in {\tt England} and one in {\tt United States}, and {\tt Yoko Ono} with one address in {\tt United States}. Note that the addresses of {\tt John Lennon} and {\tt Yoko Ono} in {\tt United States} will be matched and considered the same.

\subsection{Other Scenarios and Synopsis of Problem Formulations}
\label{sec:OtherScenariosAndSynopsis}

For reasons of space, above we have described some fundamental scenarios.
However, there are several other scenarios that can benefit
from bnode matching.
For instance, bnode matching
can be beneficial (a)
for {\em visualizing} the differences between two datasets, (b)
for  {\em query-based monitoring} of semantic warehouses
(like those constructed by  \cite{tzitzik2014matware})
by using bnode matching for comparing the answers of queries
over different versions of the warehouses,
 (c) for {\em ontology matching} (e.g. \cite{CheathamH13,NgoBT13}) by matching  the blank nodes
 that appear in schemas defined in OWL (note that blank nodes are used to represent
 class intersections, property restrictions and class axioms), and (d) leaning graphs.


Now we investigate the common axis upon which all the aforementioned tasks
tackle  bnode matching.
In brief, both the {\em equivalence} and {\em entailment} checking
perform matching as a {\em decision problem}.
In particular, they have to compute a mapping between two sets of nodes such that the triples of the one graph to be part (or subset respectively) of the other graph.
{\em $\Delta$-based} storage for versioning and
synchronization tasks perform bnode matching as an optimization problem (as formulated in \cite{bnodesDelta:2012}), because they aim at minimizing the stored (or sent respectively) delta.
It is important to note that the first problem
 can be seen as a special case of the
{\em optimization problem} where the decision will be positive
if the minimization of the delta is equal to 0 (i.e. there are no changes and the graphs are equivalent). The second problem cannot be solved through the optimization problem, as the latter computes as optimal solution(s) only injective map(s).
For the task of {\em data integration} the bnode matching optimization problem
could be applicable by introducing a less strict edit distance metric (than that described in \cite{bnodesDelta:2012}), where the named resources would not be compared as identity functions over their names (as applied in the previous tasks), but the named resources could be compared according to
URI matching techniques (as those mentioned in \S \ref{sec:URIMatching}).

\begin{figure}[h!] 	
    \vspace*{-3mm}
	\centering{
		\fbox{
        \includegraphics[scale=0.35]{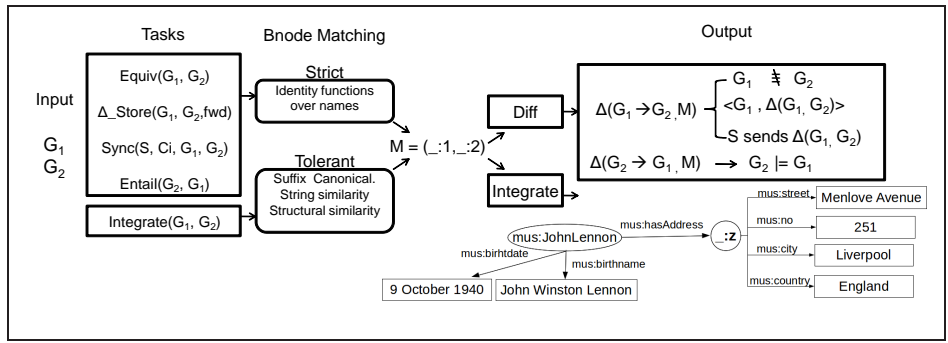}
        }
    	\caption{Summary of described tasks that require bnode matching}
		\label{fig:allTasks}
    \vspace*{-3mm}
	}
\end{figure}

To synopsize, the key point is that
we can find the sought mapping in all the above cases,
if we formulate this problem as an {\em optimization problem}.
Figure \ref{fig:allTasks} gathers all the tasks and shows the results of the tasks over the first two graphs of the running example.

\subsection{Formal Synopsis }

Below we summarize the required problems of bnode matching.
Get  $G_1$ and $G_2$ denote  two RDF graphs, and  assume that $|B_1| \leq |B_2|$.
\bi
\item[Problem \ref{problem:Equivalence}] (BM for Equivalence Testing):\\
Check if there is a bijection $M: B_1 \rightarrow B_2$
such that
 $( s, p, o )$ is in $G_1$ if and only if a triple
        $(s, p, o)\#M$ is in $G_2$. Equivalently, for $\forall$ $b_1 \in B_1$ $\exists$ $b_2 \in B_2$ such that $b_2 = M(b_1)$ and $triples(b_2) = triples(b_1)\#M$.

\item[Problem \ref{problem:Entailment} (BM for Entailment):]\  \\
Check if there is a function $m: B_2 \rightarrow B_1 \cup U_1$ such that
    for each blank node $b_2 \in B_2$,
    it holds:
    $triples(b_2)\#m \subseteq G_1$,
    where  $triples(b_2)$ is the triples in $G_2$ in which $b_2$ participates to.

\item[Problem \ref{problem:Delta} (BM for Delta-based Storage):]\ \\
Find an injection $M: B_1 \rightarrow B_2$ that minimizes the delta of the unnamed parts of the graphs
 (as  defined in \cite{bnodesDelta:2012}).
 Specifically, if $J$ denotes  all  possible injections  $B_1 \rightarrow B_2$,
 then the objective is to define the injection $M: B_1 \rightarrow B_2$ such that
$
M = arg_m(\min_{m \in J}Cost(m))
$,
where the cost of a injection $m$,
is defined as
$
Cost(m) = \sum_{b_1 \in B_1} dist_m(b_1, m(b_1))
$.
The edit distance $dist_m(b_1,b_2)$ is defined as the number of additions and deletions of triples that need to be done in order to make the triples that $b_1$ and $b_2$ participate the same, supposing that we have an injection $m$.
Specifically, and using the notations defined earlier,
  $dist_m(b_1,b_2) = |\Delta_e(triples(b_1) \rightarrow triples(b_2), m)|$.


\item[Problem \ref{prob:IntegrationFormulation} (BM for Integration):]\  \\
In comparison to
    Problem \ref{problem:Delta},
    here  during the computation of the edit distances
    we  should not consider that URIs and literals are matched as identity functions (i.e. through string equality $=$), but according to equivalence rules ($\equiv$) that are provided as input.
  Specifically, and using the notations defined earlier,
  to find the desired bnode mapping it is sufficient to solve the same
  optimization problem (as in the previous bullet),
  with the only difference that the edit distance does not consider only $m$
  but also the $\equiv$.
  In other words  the equivalence relation that is fed to the $dist$ should
  be the union $m \cup \equiv$,
  i.e. the distance between two bnodes is now quantified as:
  $dist_m(b_1,b_2) = |\Delta_e( triples(b_1) \rightarrow triples(b_2), m \cup \equiv)|$.

\ei

\vspace*{2mm}

Below we attempt to generalize and come up with a problem formulation
that can cover  the previous problems.
It is not hard to see
that Problem 1, 3 and 4
can be generalized to the following formulation:

\begin{myproblem}[Generalized BM Problem]\ \label{problem:GeneralProblem} \\
\Eisodos: Two RDF graphs $G_1, G_2$ and an equivalence relation $\equiv$ (where $\equiv \subseteq (U_1 \cup L_1) \times (U_2 \cup L_2)$). \\
\Exodos:
An injection $M: B_1 \rightarrow B_2$ such that
$
M = arg_m(\min_{m \in J}Cost(m))
$,
where $
Cost(m) = \sum_{b_1 \in B_1} dist_m(b_1, m(b_1))
$
and   $dist_m(b_1,b_2) = |\Delta_e( triples(b_1) \rightarrow triples(b_2), m \cup \equiv)|$.
\end{myproblem}

We shall write $M = Solution(G_1, G_2, M_{\equiv})$ to express that $M$
is the solution of Problem \ref{problem:GeneralProblem}.
If we solve
Problem \ref{problem:GeneralProblem},
then we have also solved
Problems 1, 3, and 4,
as we show next.
Here we focus on the Bnode Matching (BM) sub-problem of the problems and not full ones
(the latter are discussed afterwards).

\bi
\item[Pr. \ref{problem:Equivalence}]
 (BM for Equivalence Testing)

YES iff $|B_1|=|B_2|$ and $Cost(Solution(G_1, G_2, \emptyset))=0$.

Note that the first condition is required because we want a bijection.

\item[Pr. \ref{problem:Delta}]  (BM for Delta-based Storage)

The solution here is the computed injection by solving Problem 5 for  $M_{\equiv}=\emptyset$,
i.e.
$M = Solution(G_1, G_2, \emptyset)$.

\item[Pr. \ref{prob:IntegrationFormulation}]  (BM for Integration)]

The solution here is the computed injection if
we solve Problem 5 where in  $M_{\equiv}$ we provide the equivalence relation of the URIs,
i.e. \\
$M = Solution(G_1, G_2, \equiv)$.
\ei

\noindent
Let $M_{sol}$ denote the solution of
Problem \ref{problem:GeneralProblem}
for the corresponsing BM problem.
Below we show how the full problems are solved:

\bi
\item[Problem \ref{problem:Equivalence}] (Equivalence) \\
Yes, if  $\Delta_e(G_1 \rightarrow G_2, M_{sol}) = \emptyset$

\item[Problem \ref{problem:Delta}]  (Delta-based Storage). \\
The minimum in size delta is
$\Delta_e(G_1 \rightarrow G_2, M_{sol})$.
\item[ Problem \ref{prob:IntegrationFormulation}]  (Integration)] \\
If the objective is to have an integrated corpus,
i.e. a graph $G$ that contains $G_1$, $G_2$ and $M_{\equiv}$,
then  the pairs in $M_{sol}$
should also be added in $G$ as equivalence relationships,
i.e. one should add
one triple
{\tt (b, owl:sameAs, b')} for each $(b,b') \in M_{sol}$.

\ei

\noindent
As mentioned earlier,  Problem 5 does not
cover Problem \ref{problem:Entailment} (BM for Entailment).
Let $M2$ denote the function that is the solution of Problem \ref{problem:Entailment}.
The full entailment problem is then solved as follows:
\bi
\item[Problem \ref{problem:Entailment}] (Entailment)

YES iff  $\Delta_e(G_2 \rightarrow G_1, M2)$ contains only triple additions.
\ei

\noindent
The algorithms that can be used
for solving
Problem \ref{problem:GeneralProblem}
and
Problem \ref{problem:Entailment}
are discussed in Section \ref{sec:BMA}.

\subsection{Matching Named Entities versus Matching Unnamed Entities}
\label{sec:MNvsMUNshort}

With bnode matching we refer to the problem of having to define mappings
between bnodes where a bnode does not have a meaningful or persistent name.
With {\em instance matching} (else called {\em entity resolution}) \cite{ilprints859}  we refer to the problem
of having to map the same real-world {\em entities}.
There is a variety of approaches for entity resolution
(gathered in \cite{DBLP:books/daglib/0029346,Stefanidis:2014}),
most of them coming from the area of relational databases.
These works aim at matching the tuples of two tables.
This formulation is rather simplistic for the case of bnodes,
e.g. when we have connected bnodes.
In the Semantic Web literature, most works (e.g. see \cite{Dorneles,EngmannM07,noessner2010leveraging}) identify an entity as one or more named instances, which bring meaningful or persistent information,
and these works exploit lexical similarities.
The unnamed elements are in most cases neglected, and therefore
their  matching requires special treatment.
However, we should note  that
there are approaches, like collective entity matching \cite{Bhattacharya},
whose applicability, effectiveness and efficiency for bnode matching
is worth investigating.

Overall, Table \ref{tbl:BvsE}
lists a number of tasks and indicates whether
bnode and instance matching is required.
In addition  we should also note that most entity matching techniques formulate
the problem as a decision, rather than as an optimization problem.
However, note that in terms of tasks, like delta-based storage or synchronization, the bnode matching is
intrinsically an optimization problem (i.e. aims at minimizing the stored/sent information).

\begin{table}[h!]
{\scriptsize
        \begin{center}
        {\small
        \caption{Tasks requiring Bnode Matching and/or Instance Matching}
        \vspace*{0mm}
        {\small
        \btbl{|l||p{3cm}|p{3cm}|}\hline
{\em Task}  &  {\bf Bnode} Matching is required  & {\bf Instance} matching is required \\\hline
Equivalence Testing &  Yes & No \\\hline
Entailment Testing  &  Yes & No \\\hline
Delta-based storage of versions & Yes & No \\\hline
Synchronization & Yes & No \\\hline
Integration     & Yes & Yes \\\hline
\etbl
        \label{tbl:BvsE}
        }
        }
        \end{center}
        \vspace*{-2mm}
}
        \end{table}


\section{Current Situation  and Challenges}
\label{sec:DC}

At first (in \S \ref{sec:BMA}) we discuss current approaches and algorithms for
bnode mapping,
then (in \S \ref{sec:Realizations}) we discuss
where (as regards technology)  these algorithms could be realized,
and finally (in \S \ref{sec:Challenges})
we identify some challenges.\\

\subsection{Algorithms for Bnode Matching }
\label{sec:BMA}

Here we discuss algorithms that can be used (or could be used)
for solving the aforementioned problems.

\bi
\item[Problem \ref{problem:Delta}]

There are works (e.g. \cite{zdiff07,zeginisTWEB}) proposing differential functions that yield reduced in size deltas (in certain cases) but treat blank nodes as named nodes, while other works \cite{Noy2002,klein2002ova,BernersLee2004,SemVersion2005} perform a blank node matching that works under conditions and not for the general case. Such conditions are: i) when the bnodes are parts of uniquely identified triples, ii) when the bnodes have functional term labels, and iii) when the compared graphs are derived from the same system.
Jena \cite{carroll02matching} focuses only on deciding whether two KBs that contain blank nodes are equivalent or not, and do not offer any delta size saving for the case where the involved KBs are not equivalent.
To the best of our knowledge, the only work
that
attempts  to establish a blank node mapping for reducing the delta size
for the case of non equivalent KBs, and for the general case,
is
\cite{bnodesDelta:2012,LantzakiTZ12}.
In that work,
bnode matching is implemented
over the Virtuoso Sesame  API\footnote{
    \url{http://virtuoso.openlinksw.com/dataspace/doc/dav/wiki/Main/VirtSesame2Provider}
}
and
bnode matching is offered
as  an option
when someone wants to perform a diff.
The tool is publicly available\footnote{
    \url{http://www.ics.forth.gr/isl/bnodeland}
}.

\item[Problem \ref{problem:Entailment}]

The existential semantics of blank nodes make simple entailment NP-Complete.
As a solution to Problem \ref{problem:Entailment} in \cite{Pichler_drdf:entailment} authors prove that for bounded treewidth the problem can be solved in polynomial time. The proposed algorithm, {\em Polycheck} solves the problem optimally with time complexity $\OU((m^2 + mn^{2k})$, where $|G_1|=n$, $|G_2|=m$ and $treewidth(G_2)=k-1$. Other approaches relate the entailment problem with leaning graphs and identifying lean and non-lean blank nodes \cite{Pichler_drdf:entailment,j.websem365}.

\item[ Problem \ref{problem:GeneralProblem}]

Solving Problem \ref{problem:GeneralProblem}  is more complex
than solving Problem  \ref{problem:Delta},
since we have to consider  the equivalence
relation.

However an extension could easily tackle this problem.
Let's examine it in detail.
Consider the signature-based algorithm.
The only challenge lies on deciding on a common string representation of each equivalence class. Assume an equivalence class as each group of resources that are connected with {\tt owl:sameAs} relations. Note that if two or more resources are connected between them through a common resource with {\tt owl:sameAs} relations, then all of them are considered to be part of the same equivalence class. The representative resource is going to replace the occurrence of any resource of the particular equivalence class during the {\em Signature Construction} Phase.  A simple way to choose the representative resource is to select the one with the lowest capacity requirements (i.e. smaller in length string).

\ei

\subsection{Where to Realize Bnode Matching }
\label{sec:Realizations}

Bnode matching could be realized and offered in various layers:
(a) as a functionality offered by a {\em main memory} RDF/S API,
(b) as a functionality offered as an additional option
    of a Diff {\em tool/service} of RDF/S,
(c) as a functionality offered by a {\em triple store}
   (e.g. for comparing the contents of two graphspaces),
or
(d) as  an operator offered by a {\em query language}.

\subsection{Challenges}
\label{sec:Challenges}

\subsubsection{Bnode Mapping Algorithms.}

As regards bnode matching algorithms,
and in comparison to what has been described in \cite{bnodesDelta:2012},
it is worth
(a)  identifying  more special cases where the optimization problem can be solved
polynomially and devising the appropriate algorithms,
(b) designing  better and faster approximation algorithms
(the generator presented in \cite{lantzakiGenerator2014}
allows benchmarking such algorithms),
(c) devising methods and algorithms for applying bnode mapping in big datasets
that cannot fit/be loaded to one machine.
Moreover, it is worth investigating cases where
we have to construct mappings  not only between a pair of entity sets,
but amongst a family of entity sets.
Another issue is whether in an information integration context,
we should  first do entity matching and then bnode matching,
or the other way around, or should we consider both kinds of entities together.

\subsubsection{Referability and Bnodes.}

In the LOD literature
sometimes it is suggested
not to use bnodes in a dataset,
because other datasets will not be able
to add links (i.e. triples) pointing to that bnode.
Of course, if instead of a bnode id
we decide to  use a URI, then indeed linking is possible.
However,
this suggestion
does not make much (if any) sense.
By following this suggestion,
elements which are intrinsically unnamed
will become named.
In this way humans and systems
will not be able to distinguish
them from entities which are indeed named,
and this will be a burden for
integration which is the ultimate objective
of linked data.

    \section{Concluding Remarks}
    \label{sec:CR}

    This paper discussed and justified with concrete scenarios,
    why  we have to manage unnamed information elements.
    Specifically the paper described five basic tasks or scenarios
    (equivalence, entailment, differential storage, synchronization, integration)
    each having a step that requires solving a bnode matching problem.
For each scenario, the corresponding bnode matching
problem was formally defined.
Based on this analysis,
we provided more general formulations
of the problem,
which can be useful for guiding
the required technological advances.
    Subsequently, the paper discussed methods
    for realizing and offering bnode mapping functionality,
    and finally
    it identified
    various challenges  and interesting directions
    for bnode matching algorithms,
    and for pointing to unnamed elements.

\small

\subsubsection*{Acknowledgement}
This work was partly supported by
the NoE {\em APARSEN}
   (Alliance Permanent Access to the Records of Science in Europe, FP7, Proj. No 269977, 2011-2014).

\small
\bibliographystyle{plain}
\bibliography{diffBibBnodesConf}


\end{document}